\documentclass[pdflatex,sn-mathphys-num]{sn-jnl}

\usepackage{graphicx}%
\usepackage{multirow}%
\usepackage{amsmath,amssymb,amsfonts}%
\usepackage{amsthm}%
\usepackage{mathrsfs}%
\usepackage[title]{appendix}%
\usepackage{xcolor}%
\usepackage{textcomp}%
\usepackage{manyfoot}%
\usepackage{booktabs}%
\usepackage{algorithm}%
\usepackage{algorithmicx}%
\usepackage{algpseudocode}%
\usepackage{listings}%
\usepackage{comment}

\theoremstyle{thmstyleone}%
\theoremstyle{thmstyletwo}%

\theoremstyle{thmstylethree}%

\begin{document}

\title[Critical Insights about Robots for Mental Wellbeing]{Critical Insights about Robots for Mental Wellbeing}


\author*[1]{\fnm{Guy} \sur{Laban}}\email{guy.laban@cl.cam.ac.uk}
\equalcont{These authors contributed equally to this work.}

\author[1,2]{\fnm{Micol} \sur{Spitale}}\email{micol.spitale@polimi.it}
\equalcont{These authors contributed equally to this work.}

\author[1]{\fnm{Minja} \sur{Axelsson}}\email{mwa29@cam.ac.uk}

\author[1]{\fnm{Nida Itrat} \sur{Abbasi}}\email{nia22@cam.ac.uk}

\author[1]{\fnm{Hatice} \sur{Gunes}}\email{hatice.gunes@cl.cam.ac.uk}

\affil*[1]{\orgdiv{Department of Computer Science and Technology}, \orgname{University of Cambridge}, \orgaddress{\city{Cambridge}, \country{UK}}}

\affil[2]{\orgdiv{Department of Electronics, Information and Bioengineering}, \orgname{Politecnico di Milano}, \orgaddress{\city{Milan}, \country{Italy}}}


\abstract{Social robots are increasingly being explored as tools to support emotional wellbeing, particularly in non-clinical settings. Drawing on a range of empirical studies and practical deployments, this paper outlines six key insights that highlight both the opportunities and challenges in using robots to promote mental wellbeing. These include: the lack of a single, objective measure of wellbeing; the fact that robots don’t need to act as companions to be effective; the growing potential of virtual interactions; the importance of involving clinicians in the design process; the difference between one-off and long-term interactions; and the idea that adaptation and personalization are not always necessary for positive outcomes. Rather than positioning robots as replacements for human therapists, we argue that they are best understood as supportive tools that must be designed with care, grounded in evidence, and shaped by ethical and psychological considerations. Our aim is to inform future research and guide responsible, effective use of robots in mental health and wellbeing contexts.}




\maketitle

\section{Introduction}

The growing mental health crisis and increased awareness of emotional wellbeing have led to the exploration of novel technological approaches to support mental health and emotional wellbeing. Among these, social robots—automatic and semi-automatic robotic agents aimed at interacting socially with human users \cite{RefWorks:421}—have emerged as a promising tool for simulating social interactions for variety of social tasks \cite{Henschel2021}, showing potential to positively impact human wellbeing through various interventions and interactions \cite{health2024,Robinson2019}. While much of the initial work in research addressing robots for wellbeing focused on physical rehabilitation \cite{langer2021} or on support for individuals with neurodevelopmental conditions such as autism spectrum disorder (ASD) or cognitive impairments due to ageing \cite[see][]{Marchetti2022}, recent studies have expanded the scope to investigate how robots can promote and maintain mental and emotional wellbeing in everyday life across a broader range of populations \cite{Spitale2024}. Current research in social robotics for wellbeing is addressing a diverse set of clinical and non-clinical questions, exploring the potential of these agents to prevent and monitor symptom onset \cite{Post2022,Rasouli2022}, assist with the diagnosis and assessment of specific behaviours \cite[e.g.,][]{Abbasi20241,Abbasi20242,Abbasi20243,AbbasiA,Abbasi2022}, and support interventions aimed at fostering psychological resilience, reducing distress, and enhancing overall wellbeing \cite{health2024}.

These applications represent a shift in focus toward using robots to address broader aspects of emotional and psychological wellbeing, rather than targeting narrowly defined diagnoses or conditions. By helping users to cope with daily stressors, emotional events, and enhance positive experiences, these robots are being studied as potential interventions to support challenges such as stress, anxiety, and low mood. Capabilities such as delivering mindfulness and emotion regulation exercises \cite[e.g.,][]{BodalaCG21-Roman,AxelssonSG23-HRI,coping_ijsr}, facilitating positive psychology interventions \cite[e.g.,][]{axelsson2022robots,SpitaleAG23-HRI}, providing social support \cite[e.g.,][]{coping_ijsr,Laban_blt_2023}, and offering a non-judgmental presence for emotional expression \cite{share2024} are emerging as potential forms that these robots can contribute to emotional wellbeing. As the field continues to evolve rapidly, it is crucial to identify critical considerations that warrant careful examination. These include questions about appropriate robot roles and embodiment, interaction design for sensitive wellbeing contexts, ethical implications, and the importance of evidence-based approaches in the field.

Accordingly, this paper synthesizes key insights gained from multiple studies examining robots for mental wellbeing support. For the scope of this paper, "\textit{wellbeing}" is defined as a person's overall sense of health and happiness, and how they feel about their life. It includes frequent positive affect and infrequent negative affect in a way that allows for healthy functioning in daily life and relatively high life satisfaction \cite{Busseri2010,Diener1984}. Hence, wellbeing is not just the absence of clinically diagnosed mental health conditions or illness, it also includes the presence of positive experiences and emotion \cite{Feller2018}. Drawing from our collective research experiences implementing and evaluating such interventions, we identify patterns and learnings that can inform future work in this emerging domain. While not intended as a comprehensive review, we aim to highlight critical considerations and opportunities that researchers and practitioners should consider when developing robots for emotional wellbeing applications and mental health support. We pay particular attention to real-world deployment challenges, ethical considerations, methodological evaluation of such interventions, and the need to inform the design of robotic capabilities with human-centered design principles and psychological needs. Through addressing the ongoing challenges in the field and examining these insights, we hope to contribute to the development of more effective, ethical, responsible and beneficial human--robot interaction (HRI) research that can support human emotional wellbeing while acknowledging current limitations and areas requiring further research and evidence. 


\section{Insights}

Human-Robot Interaction (HRI) research in mental wellbeing is situated within the broader landscape of technological advancements in mental health support. As robotics and artificial intelligence become more integrated into everyday life, there is growing interest in their potential to assess, monitor, and enhance wellbeing. As such work comes to existence at the intersection of psychology, healthcare, and technology, it raises fundamental questions about the validity and ethical implications of robotic interventions. Unlike traditional clinical diagnostics, which rely on validated, multi-faceted assessments, HRI approaches must navigate the complexities of self-reported data and user perception (Insight 1). Additionally, the deployment of robots in wellbeing contexts must balance feasibility, effectiveness, and ethical considerations, particularly in preventing over-reliance or emotional dependence (Insight 2). By examining both physical and virtual modalities (Insight 3), researchers aim to create accessible and responsible robotic solutions that complement, rather than replace, traditional mental health practices.

The integration of social robots in wellbeing and healthcare settings presents both opportunities and challenges, particularly in designing interactions that are ethical, effective, and aligned with human needs. This highlights the importance of interdisciplinary collaboration as clinicians play a crucial role in shaping robotic interventions to ensure medical relevance, ethical integrity, and user acceptance (Insight 4). Moreover, our insights extend to the temporal dynamics of human-robot interactions (HRI), where both short-term and long-term engagements have distinct implications for mental wellbeing. While one-off interactions can serve exploratory or situational needs, sustained engagement is often necessary for meaningful therapeutic outcomes (Insight 5). Additionally, adaptation in HRI, while often assumed to enhance user experience, is not always a necessity (Insight 6); human psychological tendencies such as mentalization and social attribution can compensate for a lack of personalization. These insights underscore the need for a balanced approach to robots design — one that considers human psychological factors, ethical considerations, and practical deployment constraints to optimize social robots' role in mental wellbeing.

In the following subsections, each of these six key insights are discussed in more detail by first providing the relevant background and the challenge.

\subsection{There is no single ground truth for mental wellbeing }

\textbf{Background \& Challenge.} 

Schuller in \cite{SchullerBChapter-2015} discusses the origin of the term ground truth and its implications for fields like affective computing where labels depend on human judgment. The term ground truth has its origins in remote sensing, photogrammetry, and surveying, where it refers to actual measurements taken on-site to validate data collected through aerial or satellite imagery. In these fields, ground truth serves as the benchmark for assessing the accuracy of remotely sensed data. 

In machine learning (ML), ground truth refers to accurate, labelled data used for training and evaluating models. It represents the ``true'' output against which the predictions of the model are compared. Ground truth consists of labelled data points that a model learns from, such as annotated images in an image classification task \cite{Russakovsky2015}. Human annotators or expert systems typically provide these labels. The ground truth serves as the standard for evaluating model performance by comparing its predictions to the expected outcomes.

In the context of mental health diagnosis, professional practitioners do not rely on a single source for their assessment, but use validated metrics (see for example the OxWell study\footnote{https://oxwell.org}) to determine clinical prevalence. RCADS and SMFQ are commonly used metrics in clinical psychiatry for depression, designed to be self-reported by children and do not require administration by clinicians (see RCADS \& SMFQ\footnote{www.corc.uk.net}). The Participant Health Questionnaire (PHQ9) \cite{kroenke2001phq} and the General
Anxiety Disorder (GAD7) \cite{spitzer2006brief} questionnaire are also commonly adopted by the National Health Service (NHS) in the United Kingdom\footnote{https://www.oxfordhealth.nhs.uk/bucks-talking-therapies/professionals/nice/}
to assess depression and anxiety in adults. However, none of these would be able to provide a single and unified mental wellbeing \emph{ground truth} agreed upon by everyone involved \--- both the person self-reporting and a general practitioner (GP) or a wellbeing team member that undertake such (initial) assessments. 

In summary, defining ground truth can be challenging in domains such as emotion recognition or medical or mental health diagnosis, where labels depend on human judgment. This introduces subjectivity and inconsistencies, as different annotators may interpret the same data differently \cite{Banse96}. Additionally, the concept of ground truth in ML continues to evolve with techniques like weak supervision, semi-supervised learning, and self-supervised learning. These approaches aim to reduce the reliance on manually labelled data while preserving model accuracy, reflecting the growing need for scalable data labelling methods \cite{GuiEtAl-PAMI24}.

\textbf{Insights.} As the ultimate goal is training ML or AI models that can make autonomous predictions from either subjective responses (e.g.,
\cite{LizaEtAl2023}) or from human-robot interactive behavioural data (e.g., \cite{AbbasiSAFJG22}), the challenges discussed above also apply to the area of social robotics for wellbeing. Self-report responding, which is commonly used in HRI reseach related to wellbeing assessment and intervention,  suffers from several disadvantages such as potentially exaggerated or understated responses (e.g., due to the person feeling embarrassed) or various biases (e.g., the social desirability bias) \cite{Northrup1996}. 


While undertaking various HRI studies in the lab or in the wild for assessing or promoting wellbeing, one conservative approach could be to ensure that the population targeted initially is a non-clinical population to ensure the effectiveness and success of the robotic solution proposed. Creating a pre-study screening procedure based on PHQ9 and GAD7 questionnaires, as was done in \cite{BodalaCG21-Roman}, \cite{SpitaleAG23-HRI}, \cite{AxelssonSG23-HRI}, or parents’ reports of psychological issues in children, as was done in \cite{AbbasiSAFJG22-Roman}, could provide one solution. It is important to be vigilant about this post-study as well, even when undertaking tertile splits (low, med and high tertile) as is commonly done in clinical practice, with a consideration for clinical thresholds (e.g., high tertile with SMFQ scores of 8 and above) by using a language that is not clinical \-- i.e., where low, med and high tertile may for example refer to ‘not at risk’, ‘heading towards risk’ and ‘at risk’ categories of wellbeing. For a more detailed discussion on this see \cite{AbbasiEtAl-ISJ2022}.

When training machine learning models on human-robot interactive data, some guiding principles from the field of affective computing may be relevant, particularly the use of a "gold standard" instead of "ground truth" \cite{SchullerBChapter-2015}. While the gold standard theoretically aligns with the ground truth, it is understood to be a slightly error-prone labeling, often seen from a "bird's-eye view." Therefore, when interpreting results, it is important to recognize that the reference used is typically the gold standard, not necessarily the true ground truth. Schuller argues that this distinction has dual implications \cite{SchullerBChapter-2015}: first, models trained on affective data are likely to have some level of error; second, test results should be interpreted with caution, as what might be deemed a "classification error" could be less significant in ambiguous situations. In affective computing field, to ensure a more reliable gold standard that approximates the ground truth, multiple annotators are employed. This approach also presents an opportunity for training predictive ML models \--- they can be trained not only on the overall gold standard but also on individual annotators' data, offering valuable insights into the nuances of human judgment.

\subsection{Robot doesn't need to be a companion to improve wellbeing} 


\textbf{Background \& Challenge.} 
The HRI literature suggests that a robot can act as a \textit{companion} or as a \textit{coach} while delivering wellbeing practices. Several studies \cite{stiehl2005design, fogelson2022impact} have investigated the use of pet-like robots as touchable \textit{companions}. For example, \citet{fogelson2022impact} explored the differences in depression and loneliness for older adults with mild to moderate dementia living in residential care after the interaction with a robotic companion (dog- or cat-like). Their results showed that the elderly loneliness and depression improved via the interaction with the pet-like companion robot.
Analogously, other studies \cite{abdollahi2017pilot, jeong2023robotic} showed how robotic companions were effective in promoting mental wellbeing in home contexts. \citet{jeong2023robotic} compared the use of a robot as a companion and as a coach that delivered positive psychology exercises in homes, and their results showed that  the robotic companion (vs. the coach) was more effective in building a positive therapeutic alliance with people, enhancing participants’ wellbeing and readiness for change.
In comparison, various studies \cite{SpitaleAG23-HRI, axelsson2023robotic, calatrava2021robotic} instead have shown the effectiveness of robots as \textit{coaches} to deliver wellbeing coaching in other contexts, such as workplaces \cite{SpitaleAG23-HRI} and public cafès \cite{axelsson2023robotic}. \citet{SpitaleAG23-HRI} conducted a study involving employees of a tech company to interact with two different forms of robotic coaches that delivered positive psychology exercises, and their results showed that the robot form significantly impacts coachees' perceptions of the robotic coach in the workplace. Similarly, \citet{alves2022robot} showed how virtual robots (i.e., robot apps) can be powerful tools for mental health coaching by providing a non-judgmental space for adolescents.


\textbf{Insights.}
It is extremely important to reflect on the role of wellbeing robots in relation to their real-world deployments. A robot does not need to be a companion to be able to improve wellbeing. Jeong et al. \cite{jeong2023robotic} have shown that the companion role was preferred over the coach one in home contexts. However, deploying wellbeing robots in everyone's home is currently infeasible. 
The high cost and the lack of options in terms of robots available for generic use on the market \cite{mahdi2022survey} is a well-known problem within the HRI community and the social robotics industry \cite{Henschel2021}. 
Therefore, having a robotic companion in our homes is impractical due to the slow progress of the robotic market nowadays, i.e., the robot manufacturers may not be able to produce affordable robots in the near future. 
Alternately, we can envision to deploy robots in public spaces, schools, and workplaces. 
As indicated by our studies, these robots have been proven to be useful to function as \textit{coaches}, delivering wellbeing practices in these specific application scenarios. 
Note that, even if we design the robots as \textit{coaches}, people may perceive the robotic coach's roles differently. For instance, our past study \cite{axelsson2023robotic}, showed that participants perceived the robotic coaches having multiple functions -- such as a social entity, a guiding voice, or a focal point -- while conducting group mindfulness practice in a public cafè.

The perception of the robot's role for wellbeing is even more important when we deploy  robots that aim to interact with vulnerable populations. 
Past studies have shown how vulnerable people may become dependent on the robotic companions. For example, Laban et al. \cite{Laban_blt_2023,laban2023opening} examined self-disclosure behaviour over time towards a social robot, and how it effected people’s emotional states. They reported that participants built a relationship with the robotic companion such that they show cues of becoming somewhat dependent to it. Participants were actively seeking additional meetings and expressing a desire to continue interacting with the robot even after the experiment ended. Many participants missed the robot, and some expressed feeling sadness about ending their engagement, underscoring the robot's significant role in aiding them in coping with distress. When replicating the study with informal caregivers \cite{coping_ijsr}, Laban et al. reports experiencing a similar trend with participants showing a certain degree of attachment towards the robot. Analogously, Spitale et al. \cite{spitale2023using} reported that children who interacted with a socially assistive robot delivering linguistic activities asked the therapists \textit{``Where is the robot?"} once the study ended. 
Such findings demonstrate the risks of dependency on the robot which may cause further mental health -related concerns once the robot is removed from the participants' lives. 

Therefore, when deploying robotic solutions for vulnerable populations (e.g., elderly or children) or extremely sensitive applications such combatting loneliness or providing emotional support, HRI researchers should define and adopt safety guardrails to prevent participants from becoming emotionally attached to the robots promoting wellbeing. For example, they can limit and constrain the time participants can interact with the robot, or they can design the robot for improving wellbeing such that it is not perceived as a friend or companion but more like an assistant, a trainer, or a coach, bearing in mind that participants may still attribute other characteristics and roles to the robot (as in \cite{axelsson2023robotic})       . Recent work has also called for ``designing for exit'' in HRI, i.e., planning and facilitating exit from interaction with a robot, in order to mitigate potential harms \cite{bjorling2022designing}.


 \subsection{Embracing Virtual Modalities of HRI for Enhanced wellbeing Support}


\textbf{Background \& Challenge.} The COVID-19 pandemic emphasized the usefulness of social robots as an assistive technology supporting people's emotional wellbeing, especially during times when infection control measures required people to maintain physical distance from each other \cite{Henschel2021,Scassellati2020,Yang2020}. However, it also affected HRI research during times of social distancing as HRI researchers were limited in their ability to use laboratory facilities, reach potential target populations, and conduct in-person studies \cite{thri_covid_20,Laban_blt_2023}. The pandemic forced most individuals (including researchers) to adopt computer-mediated means of communication (CMC) \cite{Choi2021}. Accordingly, many HRI researchers studying wellbeing interactions used CMC and virtual environments to simulate verbal interactions between robots and human users \cite[e.g.,][]{Laban_blt_2023}. While the pandemic forced researchers to use virtual and CMC methods to continue with their HRI  research \cite{thri_covid_20}, it also presented a number of opportunities. While people were struggling with social isolation and minimal social interaction during the pandemic, the role of affective robots received practical meaning. Letting these robots into people's homes via their screens was a legitimate solution accepted by many users, showing the growing potential of robots in supporting wellbeing. While some previous studies claim for the moderating role of physical embodiment \cite[e.g.,][]{Heerink2007,Kiesler2008}, recent experimental studies that compared in-person interactions with mediated interactions involving social robots have reported no significant differences in participants' perception and behaviour \cite{Laban2021,Honig2020,Gittens2022}. In fact, preliminary work by Honig et al. \cite{Honig2020} suggests that the variance between mediated HRI experiments and their in-person comparable experimental designs is limited. This is in contrast to the media richness theory (MRT) \cite{mrt_1}. MRT posits that the effectiveness of a communication medium is based on its ability to transmit information in a complex and nuanced manner. Traditionally, it suggests that face-to-face interactions are the richest form of communication, due to their capacity for immediate feedback, multiple information cues, and personal focus. While conducting physical HRIs and maintaining robots' physical embodiment offer numerous benefits, mediating interventions virtually with robots has given us valuable insights and tools, which are instrumental in improving HRIs for wellbeing.

\textbf{Insights.} Laban et al. \cite{Laban_blt_2023} found that during the pandemic, participants interacting with the social robot Pepper via a video conference platform reported enhanced wellbeing and increasingly viewed the robot as social, empathic, and competent. A follow-up study with informal caregivers yielded similar findings  \cite{coping_ijsr}. These results suggest that individuals seeking emotional support, such as in social isolation during the pandemic or while coping with caregiver distress, are receptive to virtual interactions with social robots, even in the absence of physical presence. Therefore, affective robots can effectively engage and build rapport with humans to support their wellbeing, even through a medium traditionally considered ``leaner'', like video chats, relaying on context and settings \cite[see][]{RefWorks:481}. This demonstrates that when people are in need, emotional and social presence, even when mediated by technology, can potentially compensate for the lack of physical presence and nonverbal cues in HRI. Therefore, the richness of a medium in HRI may be more dynamic and context-dependent than MRT \cite{mrt_1} accounts for.

It is also important to recognise that online robots offer a logistical edge as compared with other mediums like chatbots and in-home robots in terms of availability, accessibility and ease of use. For example, firstly, online robots can employ established video conferencing platforms like Zoom \cite{Laban_blt_2023} which are frequently familiar to users across all age groups because of their prevalence in the Covid-19 pandemic\footnote{https://www.zoom.com/en/blog/90-day-security-plan-progress-report-April-22/}. Secondly, in comparison with chatbots and mental wellbeing mobile and computer applications that demand a certain level of initiative from the users\footnote{https://www.calm.com/blog/mood-monitoring} which would often be difficult to encourage especially in vulnerable population groups like children and the elderly. Finally, online robots could be made accessible using easily available smart devices, providing a monetary advantage over in-home robots that would require additional purchasing on behalf of the users\footnote{https://ddlbots.com/products/vector-robot}.

It becomes crucial for researchers to reevaluate the components essential in physical HRI geared towards enhancing wellbeing. This reevaluation should identify key elements uniquely beneficial in physical interactions. For instance, tactile feedback \cite[e.g.,][]{Matheus2022,Geva2020} or a robot's three-dimensional presence may offer irreplaceable advantages in certain therapeutic or supportive contexts. Rehabilitation robots, for example, aid individuals with neurological disorders by providing guidance, teaching physiotherapy exercises, and are crucial in enhancing independence and quality of life within their physical settings \cite[e.g.,][]{10.1145/3319502.3374797,Dembovski2022,bar-on2021}. However, in scenarios where physical elements are not central to the efficacy of the intervention, researchers should consider the advantages of online modalities. Virtual interactions provide easier access for populations in need, especially those geographically isolated or with mobility challenges. For instance, a study by Abbasi et al. \cite{Abbasi2022} examined a child-robot intervention for mental health assessment and found that virtual means allowed longitudinal replication with children from under-represented countries \cite{Abbasi20241}. Additionally, informal caregivers, often struggling to find time, space, and energy to improve their wellbeing \cite{Revenson_book1_2016}, resulting in hidden morbidity \cite{Braun2007,Sambasivam2019}, were approached via online means, providing a unique opportunity to support them and develop a much-needed robotic intervention using virtual means \cite{coping_ijsr}. Online platforms can facilitate long-term interactions, allowing continuous support in users' natural environments. This is significant in extending the reach of therapeutic interventions to wider and more diverse populations, promoting inclusivity in mental health support.

Nevertheless, it is important to note that online interactions are not meant to replace physical interactions, but rather complement them. This is especially pertinent considering the disembodied technologies and agents available. For example, Croes et al. \cite{Croes2020} showed that people were unlikely to confide with the chatbot Mizuko for extended periods, contrasting previous studies in HRI aimed at supporting wellbeing \cite[e.g.,][]{BodalaCG21-Roman,Laban_blt_2023,jeong2023robotic}. Additionally, previous study \cite{Laban2021} found that when talking to a virtual assistant (Google Tab) people's voices were more harmonious compared to when speaking with a humanoid robot, resembling the way people communicate commands at their voice assistants. In contrast, when communicating with a humanoid robot (NAO by SoftBank Robotics), people's voices were higher in pitch, possibly triggered by the robot's child-like embodiment. These findings suggest that interactions with higher-embodiment agents, such as robots with human-like design, involve more complex social perception and behaviour mechanisms. These are unconsciously manifested in responses like mimicry and imitation \cite{Chartrand1999}, underscoring the human effort to establish rapport and social relationships through visual stimuli and affordances. Therefore, the goal is not to diminish the value of physical embodiment and in-person contact but to enhance the overall experience and accessibility of interventions. By integrating both physical and virtual interactions, researchers can develop more nuanced and effective interventions tailored to specific audience needs and preferences. This dual approach can lead to more personalized, flexible, and responsive HRI experiences, ultimately contributing to the wellbeing of a broader range of individuals.

\subsection{In wellbeing settings, clinicians need to be involved in the design process}

\textbf{Background and Challenge.}
Designing social robots, especially in the context of wellbeing and mental health, should involve all relevant stakeholders, in order to integrate valuable insights throughout all phases of the development of robotic initiatives, spanning from design to deployment. For example, in healthcare settings, inputs from doctors and nurses promote the alignment of  robot functionality with real-world clinical needs and medical pipelines \cite{winkle2018social}. In addition, clinicians can provide critical insights into potential ethical concerns with regard to user privacy, so that they are addressed early at the design stage. Their involvement is key to understanding the practical, ethical and emotional considerations that affect patient outcomes, thereby, ensuring acceptance and successful use in healthcare settings \cite{winkle2018social,vsabanovic2014participatory, van2016healthcare}. Previous works examining mental wellbeing have explored including clinicians in the design process. For instance, when designing a robotic psychology coach for the workplace, Spitale et al. \cite{SpitaleAG23-HRI} chose and adapted four different positive psychology exercises to be conducted by the robot together with a professional mental wellbeing coach. They also designed the robot's personality (as expressed through its verbal behaviours, voice, and gestures) by consulting two professional wellbeing coaches. Similarly, Axelsson et al. \cite{axelsson2023robotic} designed meditation sessions for a robotic mindfulness coach together with a mindfulness instructor. To design and deploy a robot for children's mental wellbeing assessments, Abbasi et al \cite{AbbasiSAFJG22-Roman} collaborated with psychiatrists and psychologists with expertise in clinical child psychiatry and mental wellbeing assessments. As such, participatory design involving all relevant stakeholders might help in developing robots that capture valuable insights in order to promote superior mental health among the users.

\textbf{Insights.} It is important that clinicians are included in the design team of wellbeing robots throughout the design and deployment process, in order to integrate their valuable insights. One approach to integrating clinicians in the design process is human-centred design. In the past, human-centred design has been applied to the design of robots in wellbeing \cite{axelsson2021participatory, axelsson2022robots, alves2022robots}, in order to include users in the design process, and identify and prioritise users' needs and preferences in the design of robots. 
In the case of \citet{axelsson2022robots}, professional wellbeing coaches were also included in this process. These wellbeing coaches helped shape recommendations for wellbeing robots. For instance, professional coaches noted that a robot's verbal adaptation should be limited in the case of wellbeing exercises, even while users advocated for verbal adaptation \cite{axelsson2022robots}. Coaches noted that changing specific wordings of established wellbeing exercises might negatively impact their effectiveness \cite{axelsson2022robots}. This example illustrates how clinicians can enrich the human-centred robot design process, and surface directions for design that might at first glance seem unintuitive to users or roboticists. 
These professional perspectives (e.g., psychologists', coaches', and psychiatrists') should be included into the design of the robot with at least the same importance as users', particularly with a vulnerable user group such as children. wellbeing professionals should be consulted about what wellbeing practices may be appropriate for a particular context, and how those wellbeing practices should be deployed in an appropriate and safe manner (see e.g., \cite{SpitaleAG23-HRI, AbbasiSAFJG22-Roman}).

\subsection{One-off HRI may be not enough for improving mental wellbeing, but it can help!} 


\textbf{Background \& Challenge.}
Mental wellbeing coaching aims at assisting individuals who are mentally healthy in thriving in their personal and professional lives \cite{hart2001coaching} as opposed to psychological therapy which treats mental health conditions. Coaching's main goals include enhancing the coachee's hope, goal-striving and general wellbeing. 
Various coaching styles can be adopted with different timing (i.e., from a single session to several months or years) \cite{green2006cognitive} and the success of these approaches may vary from person to person. As pointed out by the human wellbeing professionals who collaborate with us in our studies, some coachees may need only a single session to acquire the tools and resources needed to achieve their goals and some others may need more time \cite{axelsson2022robots}.
For example, the \textit{miracle question} is a technique used in Solution-Focused Therapy that encourages coachees to focus on solutions rather than on problems \cite{de2012more}. It invites the coachee to envision and discuss a possible world in which their challenges have been resolved and concerns eliminated \cite{strong2009constructing}. This technique can be employed in a single session or across multiple sessions. This is due to the willingness of the coachee to achieve a specific goal in a limited time or not and the wellbeing coach capability to use the insight it provides with the \textit{miracle question} to “bring forth [coachees]’ ideas, goals, and possibilities” to reveal the potential outcomes of coaching \cite{strong2009constructing}. Some other practices, such as Positive Psychology (that helps the coachees to focus on the positive things of their lives rather than the negative ones \cite{seligman2007coaching}), may need longer practice time (e.g., from few weeks to months) to be effective. 
Therefore, when designing robots that can promote or assess mental wellbeing in individuals \cite[e.g.,][]{AbbasiSAFJG22-Roman, SpitaleAG23-HRI}, identifying the appropriate timing for the coaching practice or assessment may be challenging. This is not only because the coaching/assessment should be tailored to the specific needs of the coachees but also because the understanding of the effectiveness of such robots for wellbeing is still in its early stages. 

\textbf{Insights.} Within the HRI literature, very recent works have evaluated robots for wellbeing mostly conducting one-off studies \cite{AbbasiSAFJG22-Roman, Matheus2022} given the exploratory nature of this emerging research field. 
For example, \cite{AbbasiSAFJG22-Roman} conducted a study in which 41 children interacted with a Nao robot for a 45-minute single session to aid the evaluation of their mental wellbeing. Also, Matheus et al. \cite{Matheus2022} undertook a single-session user study with 43 participants who interacted with a novel robot \textit{Ommie} that supports deep breathing practices for the purposes of anxiety reduction. In our own study \cite{Laban2025AReappraisal} we found that participants exhibited positive trends in emotional expression during a cognitive reappraisal intervention. These trends emerged within each individual session, across a repeated longitudinal design. Participants shared more, used a greater number of adjectives to describe their experiences, and showed increased facial arousal as well as more positive facial valence as the session progressed and in response to the robot’s reappraisal suggestions. This indicates that, while long-term interaction is crucial for sustained change, each session can still offer meaningful opportunities for emotional engagement, supporting the idea that even one-off or short-term interventions can contribute positively to users’ emotional processes.

Undertaking one-off studies may be beneficial to prepare for longer deployment and may be preferable for various reasons, such as for de-risking the human-robot interaction, or for engaging in the early steps of an iterative design process in which the goal is to improve the HRI design. For instance, \citet{axelsson2025participant} conducted a one-off interaction study with a positive psychology robotic coach, in order to inform later steps of the iterative design process \cite{axelsson2022robots}. Sometimes one-off sessions can be useful to investigate short-term wellbeing applications such as public demonstrations (museums, science fairs, hospitals etc.) when the general public can try the robot and for example can learn concepts about wellbeing. For instance, \cite{rossi2020emotional} deployed a Nao robot in a Health-Vaccines Centre to understand the effects of the distraction provided by a social robot on 139 children's anxiety during a single vaccination.

In some other cases, it is extremely important that we extend the deployment of robots for wellbeing to multiple sessions and design longitudinal interactions given that mental wellbeing promotion and -- even more -- improvement need time to be effective. For example, \citet{spitale2023vita} proposed a novel framework, namely VITA, to develop autonomous robotic coaches, and they evaluated it by running a real-world longitudinal study in a tech company in which 17 employees interacted with the VITA-based robotic coach that delivered positive psychology practices. Their results showed that participants improved significantly their mental wellbeing after the 4-week coaching with the robot.
Analogously, \cite{jeong2023robotic} undertook a study in university accommodations in which 35 students interacted for 7 days with a Jibo robot that delivered positive psychology exercises. They found that students benefited from the longitudinal interaction with the robot by enhancing their mood and mental wellbeing.
On top of that, the relationship and alliance between the coach and the coachees is paramount for a successful practice \cite{de2017coaching}, and this can be achieved only with long-term interactions. Thus, the deployment of robots in long-term studies is essential for better understanding if and how robots can build a relationship with coachees \cite{spitale2023longitudinal}.

These works suggest that one-off studies can be suitable for specific wellbeing applications in which a single session is enough (e.g., reduction of children's anxiety during vaccination \cite{rossi2020emotional}) or for early exploratory studies on robots for mental wellbeing assessment \cite{AbbasiSAFJG22-Roman}, \cite{Matheus2022}, \cite{axelsson2025participant} in which we cannot rely on previous literature. However, long-term deployments are paramount for understanding the potential of using such robots in coaching practices to promote and (ideally) improve mental wellbeing in individuals.

\subsection{Adaptation is optional not a necessity }


\textbf{Background \& Challenge.} 
Adaptation and personalization in HRI can significantly shape user perceptions and engagement with robots. While initial encounters may be novel, users' acclimatisation leads to heightened expectations \cite{10.5555/3378680.3378740}. Rapport, crucial for HRI success, is fostered through adaptability, enabling a seamless integration of technology into daily life \cite{gratch_rapport_2021}. Adaptability is particularly meaningful in wellbeing contexts, allowing robots to tailor interactions to individual needs \cite[e.g.,][]{Churamani_acii_2022}. The dynamics of human wellbeing are inherently diverse and multifaceted, varying from person to person \cite{Diener1984,Steel_swbp_2008}, and accordingly, users might interact differently with robots based on their state of wellbeing \cite[e.g.,][]{laban2023opening}. Therefore, the ability to discern and respond to unique user characteristics and actions enhances the effectiveness of robots as supportive agents, addressing mental health, emotional needs, and physical requirements with nuance \cite{ahmad_adaptivity_2017,Churamani_adapt_2020}. As AI technologies advance (e.g., Generative AI and LLM), the deployment of adaptation and personalization becomes more feasible, and could be instrumental in fostering a helpful relationship between humans and robots \cite{zhang_llm_2023}, potentially amplifying the overall impact on human wellbeing. 

However, a sensible balance is needed between adaptation and standardisation. In the settings of wellbeing there are also benefits from applying a standardised approach, prioritising adherence to protocols over individual requirements, especially when evaluating the effect of such interactions with human users \cite{hoffman_primer_2020}. In certain contexts, a universal and consistent interaction model might be more suitable for ensuring predictability and reliability in the robot's behaviour \cite{akalin_safety_2023, axelsson2022robots}. Moreover, in interactions dealing sensitive content such as wellbeing, some users may prioritise privacy and prefer a more limited exchange of personal information with robots \cite{share2024}. In such cases, a less adaptive approach can mitigate concerns related to data privacy and security. Achieving this balance is crucial for the systematic understanding of a robot's impact on human wellbeing. Additionally, technological constraints \cite[see][]{Churamani_adapt_2020} and ethical considerations \cite[see][]{zhang_llm_2023,Wullenkord2020,lee_ethics_2022, axelsson2022robots} present challenges to adaptation in HRI, requiring to consider cautious and standardised approaches.  

\textbf{Insights.}
Despite the apparent absence of sophisticated adaptation and personalization in robotic behaviour, people often engage in a process of \textit{mentalization} \cite[see][]{broadbent_mental_models_1984}, attributing a sense of intention and meaning to the robot's actions \cite{Epley2007}. For example, a study found that people employ a shared set of behavioural explanations for both human and robotic agents, suggesting a natural tendency to attribute intentionality and mental states to robots, while highlighting unique expectations when interpreting and explaining robotic behaviours \cite{degraaf_hri_19}. It has been suggested that people are likely to engage in mentalization when experiencing different individual differences \cite{kilenmann_mental_2018}, such as cognitive abilities or answering to different personality traits \cite{weinstein_collabra_2022, thel2022}. A study by Spatola et al. \cite{spatola_cl_2022} showed that users experiencing higher rates of cognitive load were likely to attribute intentional properties to a robot’s behaviour. 

Other factors may arise from certain needs or a desire for meaning, wherein users recognize the benefits of interactions based on their personal needs (such as wellbeing, including low mood, loneliness, or stress). In response, they may attribute qualities to the robot accordingly \cite[see “Social Exchange Theory”,][]{homans_exchange_1958}. In other words, if a robot successfully fulfils a person's needs, the positive outcome may lead to a more favourable perception of the robot, regardless of its actual adaptability \cite{share2024}. Laban et al. \cite{Laban_blt_2023} conducted a long-term experiment showing that despite a robot's limited adaptability, participants increased self-disclosure over time, perceiving it as socially adept and reported positive mood changes. Over time, the robot became more comforting, reducing loneliness and stress. Despite a lack of adaptivity, participants found meaning in interactions, explaining that the robot seemed to get to \textit{"know them"} over time. Secondary analysis of the data reveals that introverts and individuals experiencing negative emotions tend to share more with the robot \cite{laban2023opening}. The study was replicated with a sample of informal caregivers (people who provide voluntary care to friends or family and are often challenged in coping with distress \cite[see][]{Revenson_book1_2016}). Beyond replicating previous findings \cite{Laban_blt_2023}, participants reported that despite the robot's lack of adaptivity, they were pleased to have an outlet for sharing their problems. They expressed that, \textit{"finally, something was asking them about their wellbeing and not their care-recipient"} \cite{laban_icd_2022,coping_ijsr}. Interestingly, these results align with those from similar long-term experiments with a robot that communicate in a more adaptive way by using a LLM \cite{Laban2025AReappraisal}, , with participants shifting from connection-seeking talk when distress was high to more self-development and goal-oriented themes when experiencing positive affect \cite{Laban2025l}.

Therefore, researchers should customize interactions to their population's traits and needs, balancing robots' adaptation skills with the need for standardization in wellbeing settings. When seeking adaptation in HRI interactions for wellbeing, researchers should consider human psychological tendencies, like mentalization, to support robot integration. These perceptions might stem from a deep-seated human tendency to anthropomorphize technology, or at least attribute meaning to it, imbuing it with qualities of understanding and responsiveness \cite{Epley2007,Waytz_wsh_2014}. Alternatively, even those who acknowledge the robot's lack of tailored adaptability may experience a positive subjective outcome. The fulfilment of their needs, despite the absence of highly advanced technological adaptability, seems to lead to a favourable assessment of the robot's efficacy. In these instances, HRIs for wellbeing can transcend the limitations of the machine's objective intelligence, emphasising the significance of subjective wellbeing experience in shaping perceptions of robots capability, and furthermore, whether these agents are adaptive to them during an interaction. Accordingly, researchers are encouraged to carefully assess how adaptable a robot should be and to evaluate how well it aligns with specific populations' needs and deployment contexts.

\section{Discussion}

The insights presented in this paper, drawn from HRI research studies and implementations, reveal several critical aspects of studying and using social robots for wellbeing support. While robots show promise in supporting general emotional wellbeing, our collective research experiences reveal significant considerations regarding their current capabilities and appropriate roles in mental health contexts.

Current research implementations focus primarily on wellbeing support and preventive applications -- areas where robots have demonstrated initial value, particularly in stress reduction, companionship and social engagement, and emotional support \cite{health2024,Robinson2019}. However, the gap between these research applications and formal clinical intervention remains substantial. 
A fundamental insight emerging from these six insights is that social robots are not yet prepared for formal clinical roles, despite growing interest in their therapeutic potential. This limitation extends beyond technological constraints to encompass broader considerations about therapeutic relationships \cite{Gelso1985,Gelso1994}, validation requirements \cite{Ahn2023}, and the need for professional oversight \cite{Leenes2017,Elendu2023}. 

The foundation of any successful wellbeing intervention lies in the strength of the therapeutic alliance \cite{Stubbe2018}, a construct that encompasses trust, empathy, and collaboration between the therapist and the client \cite{Gelso1985,Gelso1994}. Robots, as non-human agents, might struggle to replicate the nuanced emotional attunement and dynamic responsiveness that underpin such alliances. While some studies suggest that users may anthropomorphize robots and perceive them as empathic \cite[e.g.,][]{Laban_blt_2023,coping_ijsr,Yang2024}, this perception often reflects users’ projections rather than the robots’ actual capabilities. This gap can limit the depth of emotional connection and the perceived authenticity required for effective therapeutic outcomes. Researchers have also questioned whether forming such relationships, alliances and emotional connections with social robots is desirable or ethical in the first place \cite{de2016ethical}. Future work should focus on identifying ways to complement rather than replace the human elements of therapeutic relationships, perhaps by positioning robots as supportive tools within a human-led therapeutic framework.

Another critical point is that for social robots to move beyond experimental settings and into clinical applications, rigorous validation is critical. Clinical practices are founded on evidence-based approaches that require interventions to demonstrate not only efficacy but also safety and reliability across diverse populations \cite{Titler2008}. Current studies often rely on short-term measures and non-clinical populations, limiting the generalizability of findings, as well as their replicability \cite{Gunes2022,Spitale2023}. Longitudinal research, randomized controlled trials, and standardized assessment protocols are essential to ensure that interventions involving robots meet the stringent benchmarks of clinical practice. Researchers such as \citet{robinson2024brief} have begun taking these steps, conducting a pilot randomized controlled trial into a brief wellbeing training session delivered by a social robot, focusing on feasibility of the robot. Their work found moderate ratings for usefulness and technique enjoyment with the robot.   

Moreover, the integration of robots into wellbeing settings should call for further involvement of mental health professionals, both during the design and deployment phases. Professionals can provide critical insights into ethical considerations, ensure alignment with therapeutic goals, and establish safeguards to prevent misuse or over-reliance on robotic systems. Without professional oversight, there is a risk of inappropriate deployment, which could exacerbate vulnerabilities or lead to unintended consequences, such as dependency or reduced access to human care. This also calls for legal considerations, addressing human responsibility for robots' actions in this space \cite{Leenes2017}. Regulatory frameworks must  evolve to establish clear guidelines for the use of robots in clinical contexts, ensuring accountability and adherence to professional standards.

Therefore, our insights and discussion here emphasize the critical need for collaboration between HRI researchers and mental health professionals throughout robotic interaction and intervention development and deployment. While early research may have explored robots as potential replacements for human therapists \cite[see][]{Fiske2019}, accumulated evidence strongly indicates that successful implementation requires integration with, rather than replacement of, human practitioners. This aligns with contemporary healthcare approaches emphasizing collaborative care models \cite{Reist2022,Goodrich2013}.

The HRI studies discussed in this paper highlight important ethical considerations that emerged during research implementations. The observed potential for emotional attachment to therapeutic robots, documented across multiple studies, raises important concerns about dependency and appropriate boundaries. While this can enhance engagement and emotional support, it also introduces risks, particularly for vulnerable populations such as children, the elderly, or individuals with mental health challenges. Emotional dependency can lead to distress if the interaction with the robot is suddenly discontinued or if expectations for continued engagement are unmet \cite{Bornstein2012}. To address this, researchers and practitioners must establish clear boundaries, such as limiting interaction time and setting realistic expectations for users about the robot’s capabilities. Designing robots with roles that focus on facilitating human-human connections rather than replacing them could also mitigate the risks associated with dependency. The blurred line between therapeutic tools and perceived companions necessitates clear boundary-setting in the design and deployment of robots for wellbeing. 

Similarly, given the intimate nature of interactions often involved in therapeutic settings, ensuring robust privacy and data protection is crucial. Robots frequently collect sensitive data, including speech, facial expressions, and behavioural cues, which are meaningful for tailoring robotic interventions for wellbeing \cite{Spitale2024}. However, the storage, processing, and potential misuse of this data raise significant ethical concerns. Researchers and developers must prioritize data minimization, secure storage, and transparent communication with users about how their data is used. This is particularly important in light of the '\textit{privacy paradox}' \cite{Barnes2006} and '\textit{privacy calculus}' \cite{Culnan1999,Meier2022}, where users of robots may overlook privacy risks due to heightened perceptions of anonymity and confidentiality, as well as the additional benefits these interactions provide \cite{share2024,Laban2022}. Regulatory frameworks should suggest compliance with privacy laws, such as GDPR, and establish clear protocols for data sharing and access control.

Ethical concerns also extend to the fairness of experimental designs and deployments, as well as addressing biases in data and deployment practices. Machine learning models integrated into robots are often trained on datasets that might lack representation of diverse populations \cite{Tavares2023}, leading to unequal performance across demographic groups \cite{axelsson2022robots}. For example, models trained predominantly on Western adult data may be less effective for non-Western users, exacerbating inequities. Issues such as selection bias, unequal access to robots, and the inclusivity of research samples must be carefully considered. Studies often recruit participants from specific demographics, potentially limiting the generalisability of findings to broader populations \cite{Shea2022}. Empirical evidence of such demographic skew is already emerging; for example, a vision-language model (VLM) based pipeline for robotic wellbeing assessment of children mis-classified girls’ stories from the Children’s Apperception Test (CAT) as “clinical concern” significantly more often than boys’, despite similar true-positive rates, violating equality-of-opportunity criteria \cite{Abbasi2025vlm}. Addressing fairness requires inclusive data collection, algorithmic bias mitigation techniques, replicating studies across populations and communities \cite{Chen2023}, and aiming (as much as possible) to develop robotic interventions as affordable and scalable solutions. Researchers and robotic engineers should seek to design systems that prioritise equitable access and assess their impact across varied contexts to ensure that robotic interventions for wellbeing benefit all users, not just those with greater resources or representation.


Our research insights regarding deployment contexts suggest the need to reconsider implementation approaches for mitigating some ethical concerns. Rather than focusing on individual ownership or private therapeutic settings, experimental findings point toward potential benefits of public health approaches \cite{Cresswell2018,Willems2022}. This could include deployment in community centres, educational institutions, and workplace settings, potentially increasing access while ensuring appropriate oversight. Additionally, while much research has focused on one-on-one interactions \cite{Henschel2021}, experimental findings and on-going discussions suggest potential benefits from group and community-based applications \cite{Vanman2019, Nigro2024, selma2020, Oliveira2021}. This raises intriguing possibilities for robots as social catalysts, potentially facilitating human-human connections rather than serving solely as interaction partners. Furthermore, developing appropriate training and certification programs for mental health professionals working with robotic interventions for wellbeing can ensure these technologies are used ethically and effectively. Such programs could equip practitioners with the knowledge to address potential risks, such as dependency or privacy concerns, while optimizing the therapeutic benefits of robot-assisted interventions.

\section{Summary and Conclusion}

This paper has provided a comprehensive examination of the potential for social robots to support mental wellbeing, organized around six key insights drawn from existing research and practical implementations in HRI. It explored the nuanced challenges and opportunities of leveraging robots for wellbeing applications, offering a detailed analysis of their roles, deployment contexts, and implications for design.

The paper emphasized that the absence of a single ground truth for mental wellbeing creates challenges for assessing the efficacy of HRI interventions. Mental wellbeing is inherently multifaceted and subjective, relying on a combination of metrics, self-reports, and professional input. It also highlighted that robots need not always act as companions to support wellbeing, as roles such as coaches or facilitators can be equally effective in specific contexts, especially in shared spaces like schools and workplaces. The paper underscored the promise of virtual modalities showing how robots can extend their reach through online mediated interactions while maintaining user engagement and perceived social presence. Further, the involvement of mental health professionals in the design process was identified as critical, ensuring that robots align with therapeutic goals, address ethical considerations, and avoid unintended consequences like dependency. Longitudinal studies were emphasized as essential for understanding how relationships with robots evolve over time, with one-off studies seen as useful only for early-stage explorations. Lastly, the paper addressed the role of adaptation and personalization, arguing that while these features can enhance engagement, standardized interactions often suffice, as users attribute social and emotional qualities to robots even in the absence of advanced adaptivity. Taken together, these insights point to the significant potential of social robots in promoting mental wellbeing while acknowledging the complexities of achieving this in practice.

The insights and discussions communicated in this paper underscore that while HRI research holds promise for improving mental wellbeing, realizing this potential requires thoughtful and collaborative approaches. Robots should not be envisioned as autonomous replacements for human practitioners but as tools that complement existing mental health systems. Success in this domain hinges on careful validation of interventions, ethical consideration of user vulnerabilities, and the development of roles and interaction models that are both accessible and effective. Interdisciplinary collaboration is essential to achieve these goals. Roboticists, mental health professionals, ethicists, and policymakers must work together to address critical challenges, including scalability, data privacy, and equitable access. 

By aligning technological innovation with human-centred design and evidence-based practices, HRI researchers can contribute to the development of robots that meaningfully support mental wellbeing. These interventions must balance practicality with the ethical imperative to protect users' emotional and psychological safety. Through sustained collaboration and rigorous research, social robots can become valuable tools for promoting emotional resilience, enhancing access to support, and enriching the landscape of mental health interventions.

\section*{Acknowledgments}

G. Laban, M. Spitale, and H. Gunes have been supported by the EPSRC project ARoEQ under grant ref. EP/R030782/1. M. Axelsson is funded by the Osk. Huttunen Foundation and the EPSRC under grant EP/T517847/1. N. I. Abbasi is supported by the W.D. Armstrong Trust PhD Studentship and the Cambridge Trusts. We would like to thank Patrícia Alves-Oliveira and Kayla Matheus for their involvement in initial discussions. \textbf{Open Access:} For open access purposes, the authors have applied a Creative Commons Attribution (CC BY) licence to any Author Accepted Manuscript version arising.

\bibliography{ref}

\end{document}